# Directed hypergraph neural network

Loc Tran
Faculty of Engineering and Technology
Thu Dau Mot University
E-mail: tran0398@umn.edu
Linh Tran
Faculty of Engineering and Technology
Thu Dau Mot University
E-mail: linhtran.cntt@tdmu.edu.vn

Abstract: To deal with irregular data structure, graph convolution neural networks have been developed by a lot of data scientists. However, data scientists just have concentrated primarily on developing deep neural network method for un-directed graph. In this paper, we will present the novel neural network method for directed hypergraph. In the other words, we will develop not only the novel directed hypergraph neural network method but also the novel directed hypergraph based semi-supervised learning method. These methods are employed to solve the node classification task. The two datasets that are used in the experiments are the cora and the citeseer datasets. Among the classic directed graph based semi-supervised learning method, the novel directed hypergraph based semi-supervised learning method, the novel directed hypergraph neural network method that are utilized to solve this node classification task, we recognize that the novel directed hypergraph neural network achieves the highest accuracies.

## 1. Introduction

In recent years, deep convolution neural networks have gained much interests from data scientists and have utilized to solve many classification tasks such as image recognition [9] and speech recognition [6], to name a few. In order to deal with irregular data structure, graph convolution neural networks have been developed by a lot of data scientists such as Thomas Kipf [7]. There are two classes of graph convolution neural network. The first class of graph convolution neural network is the spatial based approach. This spatial based approach implements the convolution on the graph by accumulating information of the neighbor nodes. The second class of graph convolution neural network is the spectral based approach. This spectral based approach implements a variant of graph convolution neural network based on different graph Laplacians. Easily, we recognize that the time complexity of spectral based approach are much higher than the time complexity of spatial based approach; however, the accuracy of the spectral based approach is higher than the accuracy of the spatial based approach. In this graph data structure, we easily see that the edge carry no information about the direction. Moreover, in this graph data structure, the edge can connect only two vertices. In the other words, data scientists have concentrated primarily on developing deep neural network method for un-directed graph.

In order to overcome these two information losses which are “the edge carry no information about the direction” and “the edge can connect only two vertices” of the graph data structure which “can” affect the performance of the “node clustering task” or the “node classification task”, we employ the directed hypergraph data structure [4, 5] and develop the deep neural network method based on this directed hypergraph data structure. This method is called the spectral directed hypergraph neural network method. The development of this directed hypergraph neural network is considered the very hard task and the novel work. First, we need to define the transition probability matrix of the random walk on the directed hypergraph. Then, we need to compute the PageRank vector of the directed hypergraph. Finally, we can compute the directed hypergraph Laplacian. From the directed hypergraph Laplacian, we can start developing the spectral directed hypergraph neural network and applying this novel method to solve the classification task. The two citation datasets that are used in the classification task are the cora and the citeseer datasets. To the best of our knowledge, this research work has not been developed before.

In this paper, our contributions are three folds:

- Develop the novel directed hypergraph semi-supervised learning method.
- Develop the novel directed hypergraph neural network.

- The accuracies of the classic directed graph semi-supervised learning method (which is the baseline method), the novel directed hypergraph semi-supervised learning method, the novel directed hypergraph neural network are computed and compared.

We will organize the paper as follows: Section 2 will present the preliminary notations and definitions. Section 3 will introduce the novel directed hypergraph semi-supervised learning method. Section 4 will present the directed hypergraph neural network. Section 5 will describe the datasets and present the experimental results. Section 6 will conclude this paper and the future direction of researches will be discussed.

## 2. Preliminary notations and definitions

Given the directed hypergraph $H = (V, E)$ where $V$ is the set of vertices and $E$ is the set of hyper-arcs. Each hyper-arc $e \in E$ is written as $e = (e^{Tail}, e^{Head})$. The vertices of $e$ are denoted by $\boldsymbol{e} = e^{Tail} \cup e^{Head}$. $e^{Tail}$ is called the tail of the hyper-arc $e$ and $e^{Head}$ is called the head of the hyper-arc $e$. Please note that $e^{Tail} \neq \emptyset, e^{Head} \neq \emptyset, e^{Tail} \cap e^{Head} = \emptyset$.

The directed hypergraph $H = (V, E)$ can be represented by two incidence matrices $H^{Tail}$ and $H^{Head}$.

These two incidence matrices $H^{Tail}$ and $H^{Head}$ can be defined as follows

$$h^{Tail}(v, e^{Tail}) = \begin{cases} 1 \; if \; v \in e^{Tail} \\ 0 \; otherwise \end{cases} (1)$$

$$h^{Head}(v, e^{Head}) = \begin{cases} 1 \; if \; v \in e^{Head} \\ 0 \; otherwise \end{cases} (2)$$

The example of the directed hypergraph is illustrated in Figure 1:

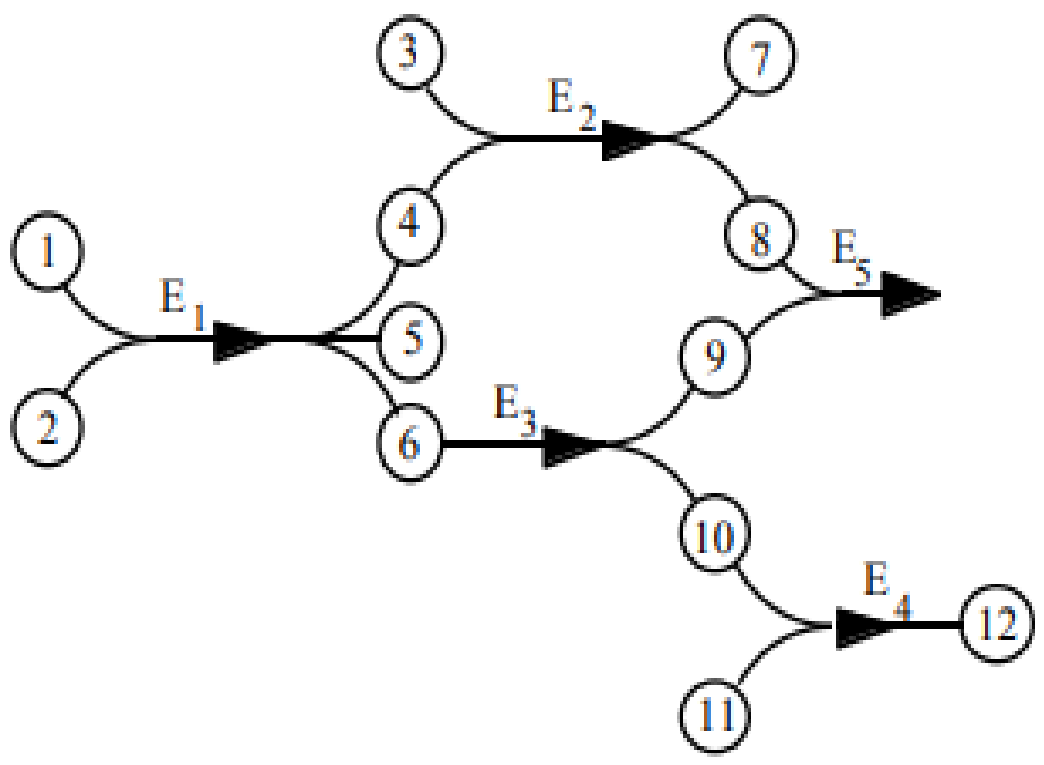

Figure 1. Directed hypergraph example with 12 vertices and 5 hyper-arcs [4].

Let $w(e)$ be the weight of the hyper-arc $e$. Let $W$ be the diagonal matrix containing the weights of hyper-arcs in its diagonal entries.

From the above definitions, we can define the tail and head degrees of the vertex v and the tail and head degrees of the hyper-arc e as follows

$$d^{Tail}(v) = \sum_{e \in E} w(e) h^{Tail}(v, e^{Tail}) (3)$$

$$d^{Head}(v) = \sum_{e \in E} w(e) h^{Head}(v, e^{Head}) (4)$$

$$d^{Tail}(e) = \sum_{v \in V} h^{Tail}(v, e^{Tail}) (5)$$

$$d^{Head}(e) = \sum_{v \in V} h^{Head}(v, e^{Head}) (6)$$

Let $D_v^{Tail}$, $D_v^{Head}$, $D_e^{Tail}$, and $D_e^{Head}$ be four diagonal matrices containing the tail and head degrees of vertices and the tail and head degrees of hyper-arcs in their diagonal entries respectively. Please note that $D_v^{Head}$ and $D_v^{Tail}$ are the $R^{|V|*|V|}$ matrices and $D_e^{Head}$ and $D_e^{Tail}$ are the $R^{|E|*|E|}$ matrices.

From [2], we know that the transition probability of the random walk on directed hypergraph can be defined as follows

$$p(u, v) = \sum_{e \in E} w(e) \frac{h^{Tail}(u, e^{Tail})}{d^{Tail}(u)} \frac{h^{Head}(v, e^{Head})}{d^{Head}(e)} (7)$$

From the above definition, the transition probability matrix $P$ of the random walk on the directed hypergraph can be defined in the matrix form as follows

$$P = D_v^{Tail^{-1}} H^{Tail} W D_e^{Head^{-1}} H^{Head^T} (8)$$

The PageRank vector $\pi$ of the directed hypergraph is the solution of the following equation

$$\pi^T = \pi^T P \quad (9)$$

Moreover, we know that the above equation can easily be solved by the Power method.

Next, we give two novel definitions of the directed hypergraph Laplacian which are un-normalized directed hypergraph Laplacian and symmetric normalized directed hypergraph Laplacian.

Let $S$ be the diagonal matrix containing all elements of PageRank vector $\pi$ of the directed hypergraph in its diagonal entries.

The un-normalized directed hypergraph Laplacian can be defined as follows

$$L = S - \frac{S {D_v^{Tail}}^{-1} H^{Tail} W {D_e^{Head}}^{-1} {H^{Head}}^{T} + ({D_v^{Tail}}^{-1} H^{Tail} W {D_e^{Head}}^{-1} {H^{Head}}^{T})^T S}{2} \quad (10)$$

The symmetric normalized directed hypergraph Laplacian can be defined as follows

$$L_{sym} = I - \frac{S^{\frac{1}{2}} {D_v^{Tail}}^{-1} H^{Tail} W {D_e^{Head}}^{-1} {H^{Head}}^{T} S^{-\frac{1}{2}} + S^{-\frac{1}{2}} ({D_v^{Tail}}^{-1} H^{Tail} W {D_e^{Head}}^{-1} {H^{Head}}^{T})^T S^{\frac{1}{2}}}{2} \quad (11)$$

From these two above definitions, we can develop the directed hypergraph Laplacian based semi-supervised learning algorithms and the directed hypergraph neural network.

3. Directed hypergraph semi-supervised learning

Given a set of samples $\{x_1, \dots, x_l, x_{l+1}, \dots, x_{l+u}\}$ where $n = |V| = l + u$ is the total number of samples (i.e. vertices) in the directed hypergraph *H=(V,E)*.

The method constructing the incidence matrix $H^{Tail}$ and $H^{Head}$ from the datsets will be specified clearly in the Experiments and Results section (i.e. Section 5).

Define $C$ be the number of classes. Please note that $\{x_1, \dots, x_l\}$ is the set of all labeled points and $\{x_{l+1}, \dots, x_{l+u}\}$ is the set of all un-labeled points.

Let $Y \in R^{|V|*C}$ the initial label matrix for *n* samples in the directed hypergraph *H* be defined as follows

$$Y_{ij} = \begin{cases} 1 \; if \; x_i \in \; class \; j \; and \; 1 \le i \le l \\ -1 \; if \; x_i \; \notin \; class \; j \; and \; 1 \le i \le l \\ 0 \; if \; l+1 \le i \le n \end{cases} \quad (12)$$

Let the matrix $F \in R^{|V|*C}$ be the estimated label matrix for the set of samples $\{x_1, \dots, x_l, x_{l+1}, \dots, x_{l+u}\}$, where the point $x_i$ is labeled as sign($F_{ij}$) for each class $j$ ($1 \le j \le C$)

Our objective is to predict the labels of the un-labeled points $x_{l+1}, \dots, x_{l+u}$. In the other words, we need to compute the final solution matrix *F*.

In short, we would like to solve the following optimization problem

$$E(f) = \frac{1}{2} \sum_{(u,v) \subseteq e} \pi(u) \sum_{e \in E} w(e) \frac{h^{Tail}(u, e^{Tail})}{d^{Tail}(u)} \frac{h^{Head}(v, e^{Head})}{d^{Head}(e)} \left(\frac{f(u)}{\sqrt{\pi(u)}} - \frac{f(v)}{\sqrt{\pi(v)}}\right)^2 + \mu ||f - y||^2 \quad (13)$$

We know that

$$\frac{1}{2} \sum_{(u,v) \subseteq e} \pi(u) \sum_{e \in E} w(e) \frac{h^{Tail}(u, e^{Tail})}{d^{Tail}(u)} \frac{h^{Head}(v, e^{Head})}{d^{Head}(e)} \left(\frac{f(u)}{\sqrt{\pi(u)}} - \frac{f(v)}{\sqrt{\pi(v)}}\right)^2$$

$$= \frac{1}{2} \sum_{e \in E} \frac{1}{2} \sum_{v \in V} \{ \sum_{u \to v} \pi(u) w(e) \frac{h^{Tail}(u, e^{Tail})}{d^{Tail}(u)} \frac{h^{Head}(v, e^{Head})}{d^{Head}(e)} \left(\frac{f(u)}{\sqrt{\pi(u)}} - \frac{f(v)}{\sqrt{\pi(v)}}\right)^2$$
$$+ \sum_{u \leftarrow v} \pi(v) w(e) \frac{h^{Tail}(v, e^{Tail})}{d^{Tail}(v)} \frac{h^{Head}(u, e^{Head})}{d^{Head}(e)} \left(\frac{f(v)}{\sqrt{\pi(v)}} - \frac{f(u)}{\sqrt{\pi(u)}}\right)^2 \}$$

$$
\begin{aligned}
&= \frac{1}{2}\sum_{e\in E}(\frac{1}{2}\sum_{v\in V}\left\{\sum_{u\to v} w(e)\frac{h^{Tail}(u,e^{Tail})}{d^{Tail}(u)}\frac{h^{Head}(v,e^{Head})}{d^{Head}(e)}f^2(u)\right.\\
&+ \sum_{u\to v} w(e)\frac{h^{Tail}(u,e^{Tail})}{d^{Tail}(u)}\frac{h^{Head}(v,e^{Head})}{d^{Head}(e)}\frac{\pi(u)}{\pi(v)}f^2(v)\\
&\left.- 2\sum_{u\to v}\pi(u)w(e)\frac{h^{Tail}(u,e^{Tail})}{d^{Tail}(u)}\frac{h^{Head}(v,e^{Head})}{d^{Head}(e)}\frac{1}{\sqrt{\pi(u)}\sqrt{\pi(v)}}f(u)f(v)\right\}\\
&+ \frac{1}{2}\sum_{v\in V}\{\sum_{u\leftarrow v} w(e)\frac{h^{Tail}(v,e^{Tail})}{d^{Tail}(v)}\frac{h^{Head}(u,e^{Head})}{d^{Head}(e)}f^2(v)\\
&+ \sum_{u\leftarrow v} w(e)\frac{h^{Tail}(v,e^{Tail})}{d^{Tail}(v)}\frac{h^{Head}(u,e^{Head})}{d^{Head}(e)}\frac{\pi(v)}{\pi(u)}f^2(u)\\
&- 2\sum_{u\to v}\pi(v)w(e)\frac{h^{Tail}(v,e^{Tail})}{d^{Tail}(v)}\frac{h^{Head}(u,e^{Head})}{d^{Head}(e)}\frac{1}{\sqrt{\pi(u)}\sqrt{\pi(v)}}f(u)f(v)\})
\end{aligned}
$$

Moreover, we know that

$$
\begin{aligned}
&\sum_{e\in E}\sum_{(u,v)\subseteq e} w(e)\frac{h^{Tail}(u,e^{Tail})}{d^{Tail}(u)}\frac{h^{Head}(v,e^{Head})}{d^{Head}(e)}f^2(u)\\
&= \sum_{u\in V}\sum_{v\leftarrow u}\sum_{e\in E} w(e)\frac{h^{Tail}(u,e^{Tail})}{d^{Tail}(u)}\frac{h^{Head}(v,e^{Head})}{d^{Head}(e)}f^2(u) = \sum_{u\in V}(\sum_{v\leftarrow u} p(u,v))f^2(u)\\
&= \sum_{u\in V} f^2(u) = \sum_{v\in V} f^2(v)
\end{aligned}
$$

$$
\begin{aligned}
&\sum_{e\in E}\sum_{(u,v)\subseteq e} w(e)\frac{h^{Tail}(u,e^{Tail})}{d^{Tail}(u)}\frac{h^{Head}(v,e^{Head})}{d^{Head}(e)}\frac{\pi(u)}{\pi(v)}f^2(v)\\
&= \sum_{v\in V}\sum_{u\to v}\sum_{e\in E} w(e)\frac{h^{Tail}(u,e^{Tail})}{d^{Tail}(u)}\frac{h^{Head}(v,e^{Head})}{d^{Head}(e)}\frac{\pi(u)}{\pi(v)}f^2(v) = \sum_{v\in V}\left(\sum_{u\to v}\frac{p(u,v)\pi(u)}{\pi(v)}\right)f^2(v)\\
&= \sum_{v\in V} f^2(v)
\end{aligned}
$$

$$
\begin{aligned}
&\sum_{e\in E}\sum_{(u,v)\subseteq e}\pi(u)w(e)\frac{h^{Tail}(u,e^{Tail})}{d^{Tail}(u)}\frac{h^{Head}(v,e^{Head})}{d^{Head}(e)}\frac{1}{\sqrt{\pi(u)}\sqrt{\pi(v)}}f(u)f(v)\\
&= \sum_{v\in V}\sum_{u\to v}\sum_{e\in E} w(e)\frac{h^{Tail}(u,e^{Tail})}{d^{Tail}(u)}\frac{h^{Head}(v,e^{Head})}{d^{Head}(e)}\frac{\sqrt{\pi(u)}}{\sqrt{\pi(v)}}f(u)f(v)
\end{aligned}
$$

$$
\begin{aligned}
&\sum_{e\in E}\sum_{(v,u)\subseteq e} w(e)\frac{h^{Tail}(v,e^{Tail})}{d^{Tail}(v)}\frac{h^{Head}(u,e^{Head})}{d^{Head}(e)}f^2(v)\\
&= \sum_{v\in V}\sum_{u\leftarrow v}\sum_{e\in E} w(e)\frac{h^{Tail}(v,e^{Tail})}{d^{Tail}(v)}\frac{h^{Head}(u,e^{Head})}{d^{Head}(e)}f^2(v) = \sum_{v\in V}(\sum_{u\leftarrow v} p(v,u))f^2(v) =\\
&= \sum_{v\in V} f^2(v)
\end{aligned}
$$

$$\sum_{e\in E}\sum_{(v,u)\subseteq e} w(e)\frac{h^{Tail}(v,e^{Tail})}{d^{Tail}(v)}\frac{h^{Head}(u,e^{Head})}{d^{Head}(e)}\frac{\pi(v)}{\pi(u)}f^2(u)$$

$$= \sum_{u\in V}\sum_{v\to u}\sum_{e\in E} w(e)\frac{h^{Tail}(v,e^{Tail})}{d^{Tail}(v)}\frac{h^{Head}(u,e^{Head})}{d^{Head}(e)}\frac{\pi(v)}{\pi(u)}f^2(u) = \sum_{u\in V}\left(\sum_{v\to u}\frac{p(v,u)\pi(v)}{\pi(u)}\right)f^2(u)$$

$$= \sum_{v\in V} f^2(v)$$

$$\sum_{e\in E}\sum_{(v,u)\subseteq e} \pi(v)w(e)\frac{h^{Tail}(v,e^{Tail})}{d^{Tail}(v)}\frac{h^{Head}(u,e^{Head})}{d^{Head}(e)}\frac{1}{\sqrt{\pi(u)}\sqrt{\pi(v)}}f(u)f(v)$$

$$= \sum_{v\in V}\sum_{u\leftarrow v}\sum_{e\in E} w(e)\frac{h^{Tail}(v,e^{Tail})}{d^{Tail}(v)}\frac{h^{Head}(u,e^{Head})}{d^{Head}(e)}\frac{\sqrt{\pi(v)}}{\sqrt{\pi(u)}}f(u)f(v)$$

Thus, we can conclude that

$$\frac{1}{2}\sum_{(u,v)\subseteq e}\pi(u)\sum_{e\in E} w(e)\frac{h^{Tail}(u,e^{Tail})}{d^{Tail}(u)}\frac{h^{Head}(v,e^{Head})}{d^{Head}(e)}\left(\frac{f(u)}{\sqrt{\pi(u)}}-\frac{f(v)}{\sqrt{\pi(v)}}\right)^2$$

$$= \sum_{v\in V}\{f^2(v) - \frac{1}{2}(\sum_{u\to v}\sum_{e\in E} w(e)\frac{h^{Tail}(u,e^{Tail})}{d^{Tail}(u)}\frac{h^{Head}(v,e^{Head})}{d^{Head}(e)}\frac{\sqrt{\pi(u)}}{\sqrt{\pi(v)}}f(u)f(v)$$

$$+ \sum_{u\leftarrow v}\sum_{e\in E} w(e)\frac{h^{Tail}(v,e^{Tail})}{d^{Tail}(v)}\frac{h^{Head}(u,e^{Head})}{d^{Head}(e)}\frac{\sqrt{\pi(v)}}{\sqrt{\pi(u)}}f(u)f(v))\}$$

Finally, in general, the closed form solution of the directed hypergraph based semi-supervised learning method can be computed as follows

$$F = (1-\alpha)\left(I - \alpha\frac{S^{\frac{1}{2}}D_v^{Tail^{-1}}H^{Tail}WD_e^{Head^{-1}}H^{Head^T}S^{-\frac{1}{2}} + S^{-\frac{1}{2}}(D_v^{Tail^{-1}}H^{Tail}WD_e^{Head^{-1}}H^{Head^T})^T S^{\frac{1}{2}}}{2}\right)^{-1}Y \quad (14),$$

where α is the parameter.

4. Directed hypergraph neural network

Let $T = \frac{S^{\frac{1}{2}}D_v^{Tail^{-1}}H^{Tail}WD_e^{Head^{-1}}H^{Head^T}S^{-\frac{1}{2}} + S^{-\frac{1}{2}}(D_v^{Tail^{-1}}H^{Tail}WD_e^{Head^{-1}}H^{Head^T})^T S^{\frac{1}{2}}}{2}$ (15)

From [7, 1, 3], the output of the directed hypergraph neural network can be defined and computed as follows

$$Z = softmax(TReLU(TX\theta^1)\theta^2) \quad (16)$$

Please note that $X \in R^{n*L_1}$ is the feature matrix (i.e. the image dataset). $\theta^1 \in R^{L_1*L_2}$ and $\theta^2 \in R^{L_2*C}$ are two parameter matrices that are needed to be learned during the training process.

We can easilly recognize that instead of adding a self loop to each node in renormalization phase as in [7], we directly use the term $\frac{S^{\frac{1}{2}}D_v^{Tail^{-1}}H^{Tail}WD_e^{Head^{-1}}H^{Head^T}S^{-\frac{1}{2}} + S^{-\frac{1}{2}}(D_v^{Tail^{-1}}H^{Tail}WD_e^{Head^{-1}}H^{Head^T})^T S^{\frac{1}{2}}}{2}$ (i.e. $T$) in the directed hypergraph Laplacian to compute the output Z of the directed hypergraph neural network. Obviously, we see that $T$ has the eigenvalues in the range $[-1,1]$.

5. Experiments and Results

In this section, we will apply the classic directed graph based semi-supervised learning method [10], the novel directed hypergraph based semi-supervised learning method, the novel directed hypergraph neural network method to solve the classification problem. The two citation datasets that we will use in the experiments are the cora dataset and the citeseer dataset [8].

### 5.1 Datasets

Cora: This dataset consists of 2,708 scientific publications classified into one of seven classes which are Case_Based, Genetic_Algorithms, Neural_Networks, Probabilistic_Methods, Reinforce_Learning, Rule_Learning, Theory. The citation network consists of 5,429 links. In the other words, this cora citation network contains 2,708 nodes (i.e. scientific publications) and 5,429 edges (i.e. citation links). For training, we use 20 samples per class. In the other words, there are 140 samples in the training set. Each publication in this cora dataset is described by a 0/1-valued word vector indicating the absence/presence of the corresponding word from the dictionary. The dictionary contains 1,433 unique words. In the other words, we have the $R^{2,708*1,433}$ feature matrix.

Citeseer: This dataset consists of 3,312 scientific publications classified into one of six classes which are Agents, AI, DB, IR, ML, and HCI. The citation network consists of 4,732 links. In the other words, this citeseer citation network contains 3,312 nodes (i.e. scientific publications) and 4,732 edges (i.e. citation links). For training, we use 70 samples per class. In the other words, there are 420 samples in the training set. Each publication in the citeseer dataset is described by a 0/1-valued word vector indicating the absence/presence of the corresponding word from the dictionary. The dictionary contains 3,703 unique words. In the other words, we have the $R^{3,312*3,703}$ feature matrix.

### 5.2 Experiments and Results

In this section, initially, we will show how to construct the directed graph from the citation network. If there is a link from node *i* to node *j*, we will construct a link from node *j* to node *i*. In the other words, there are two links (with directions) between the cited scientific publication and the citing publication. Not only the cited scientific publication influences the citing publication, but the citing scientific publication also influences the cited publication.

Next, we will discuss how to construct a directed hypergraph. First, please note that the number of hyper-arcs in the hypergraph is equal to the number of scientific publications in the dataset. There are two classes of directed hypergraph that we will construct (from the directed graph describing above) in this paper. The first class is the F-directed hypergraph. The F-directed hypergraph is the directed hypergraph whose hyperarcs are F-arcs. The F-arc is the hyper-arc whose the tail has only one node. The second class is the B-directed hypergraph. The B-directed hypergraph is the directed hypergraph whose hyperarcs are B-arcs. The B-arc is the hyper-arc whose the head has only one node. After constructing the F-directed hypergraph or the B-directed hypergraph, we will construct the two incidence matrices $H^{Tail}$ and $H^{Head}$ of the directed hypergraph. From the computed $H^{Tail}$ and $H^{Head}$, we can compute the term $\frac{S^{\frac{1}{2}}D_v^{Tail^{-1}}H^{Tail}WD_e^{Head^{-1}}H^{Head^T}S^{-\frac{1}{2}}+S^{-\frac{1}{2}}(D_v^{Tail^{-1}}H^{Tail}WD_e^{Head^{-1}}H^{Head^T})^T S^{\frac{1}{2}}}{2}$ in the symmetric normalized directed hypergraph Laplacian that will be used in the novel directed hypergraph based semi-supervised learning method and the novel directed hypergraph neural network method.

The example of the B-arc and F-arc are illustrated in Figure 2:

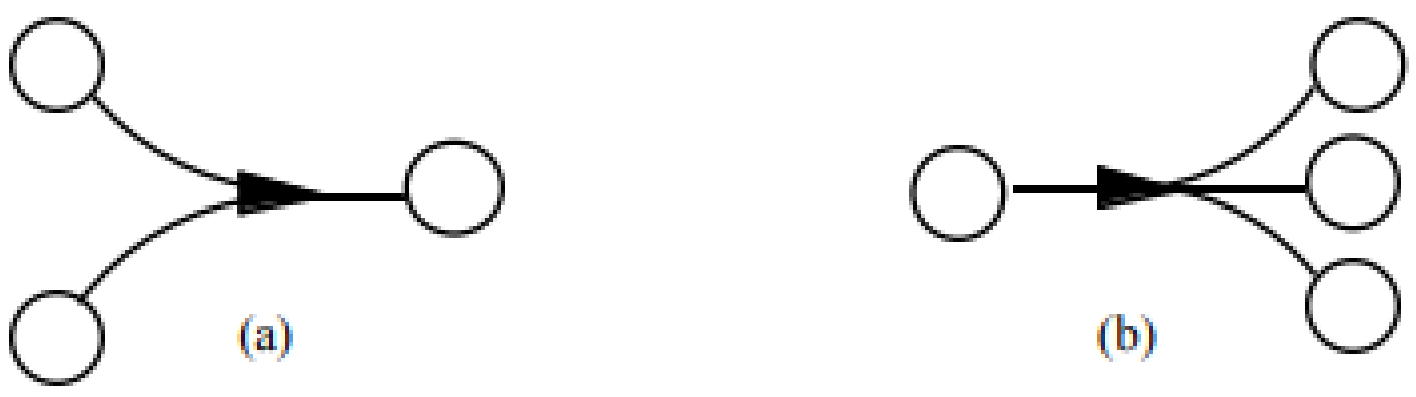

Figure 2. B-arc (a) and F-arc (b) examples [4].

We run our six methods which are the classic directed graph based semi-supervised learning method, the novel F-directed hypergraph based semi-supervised learning method, the B-directed hypergraph based semi-supervised learning method, the directed graph neural network method, the F-directed hypergraph neural network method, the B-directed hypergraph neural network method (Python code) on Google Colab with NVIDIA Tesla K80 GPU and 12 GB RAM. The following table 1 (figure 3) and table 2 (figure 4) show the experimental results of our six methods.

Table 1. Cora dataset: Comparison of our six methods. The classification accuracy (%) is reported

| Methods | Accuracy (%) |
|---|---|
| Directed graph based semi-supervised learning | 67.25 |
| F-directed hypergraph based semi-supervised learning | 67.25 |
| B-directed hypergraph based semi-supervised learning method | 67.25 |
| Directed graph neural network method | 81.42 |
| F-directed hypergraph neural network method | **81.93** |
| B-directed hypergraph neural network method | 81.85 |

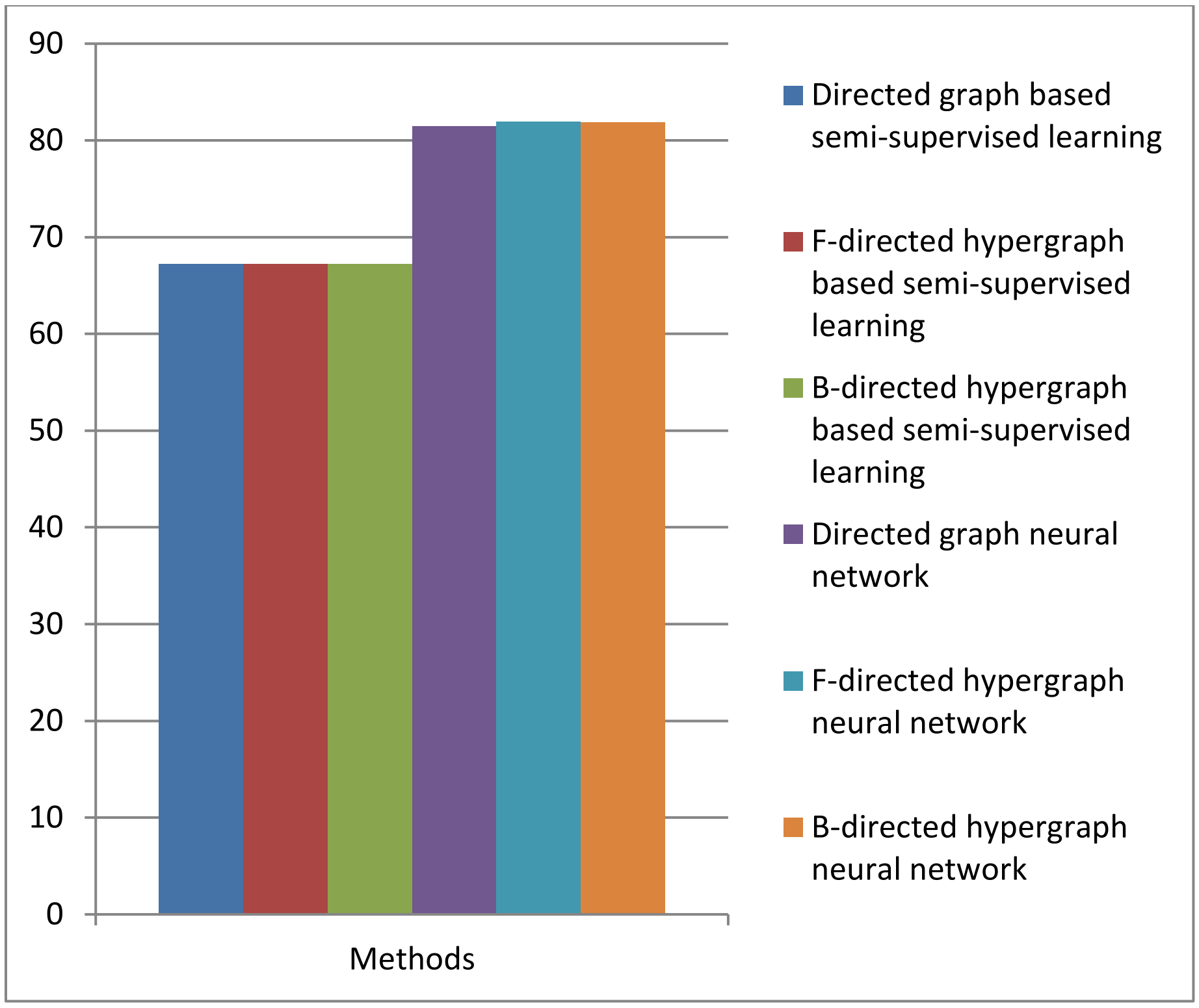

Figure 3. Cora dataset: Comparison of our six methods

Table 2. Citeseer dataset: Comparison of our six methods. The classification accuracy (%) is reported

| Methods | Accuracy (%) |
| --- | --- |
| Directed graph based semi-supervised learning | 48.23 |
| F-directed hypergraph based semi-supervised learning | 48.23 |
| B-directed hypergraph based semi-supervised learning method | 48.23 |
| Directed graph neural network method | 69.84 |
| F-directed hypergraph neural network method | **70.53** |
| B-directed hypergraph neural network method | 70.43 |

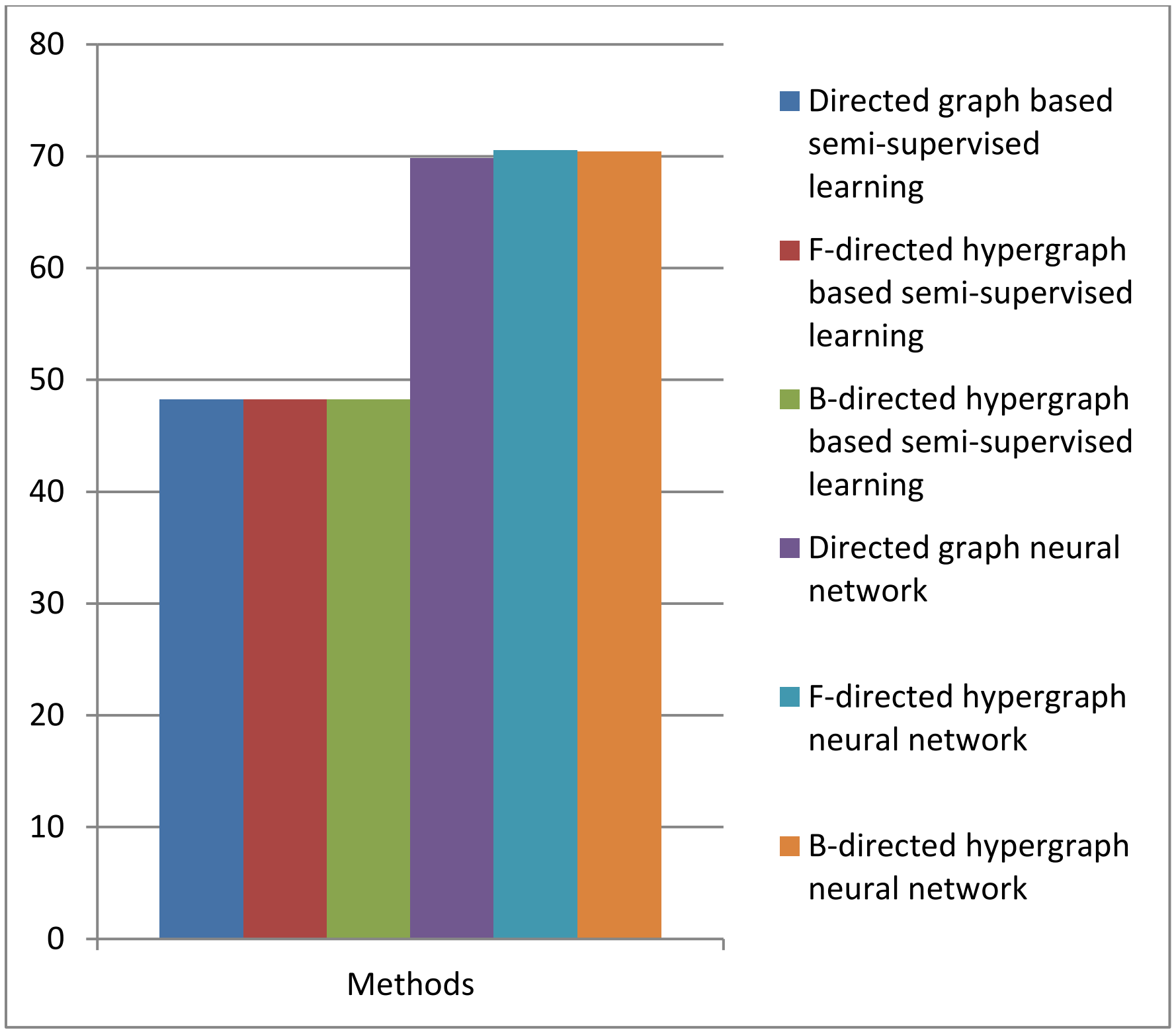

Figure 4. Citeseer dataset: Comparison of our six methods

From the above tables and figures, we easily see that the directed graph neural network method and the directed hypergraph neural network method significantly are better than the classic directed graph semi-supervised learning method and the novel directed hypergraph semi-supervised learning method. Moreover, the F-directed hypergraph neural network method and the B-directed hypergraph neural network method are slightly better than the directed graph neural network method. In general, for these two cora dataset and citeseer dataset, the F-directed hypergraph neural network method reaches the highest accuracy performance measures.

Last but not least, interestingly, the novel directed hypergraph based semi-supervised learning method does not outperform the classic directed graph based semi-supervised learning method. We think that the way constructing the directed hypergraph from the directed graph is not good enough. In the future, we will investigate more carefully how to construct the directed hypergraph from the directed graph.

## 6. Conclusions

In this paper, we have successfully developed the novel directed hypergraph based semi-supervised learning method and the novel directed hypergraph neural network method. Experimental results show that the directed hypergraph neural network method significantly outperforms the novel directed hypergraph based semi-supervised learning method and the classic directed graph based semi-supervised learning. Moreover, the F-directed hypergraph neural network method achieves the highest accuracy performance measures for the two datasets: cora and citeseer among other directed (hyper)-graph based methods.

Last but not least, in the future work, we will combine the directed hypergraph p-Laplacian based semi-supervised learning method with the directed hypergraph neural network to construct a novel directed hypergraph neural network method. Finally, we can apply this novel method to various datasets such cora, citeseer, and pubmed, to name a few.